\documentclass[12pt]{iopart}

\usepackage{cite}
\usepackage{amssymb,amsfonts}
\usepackage{algorithmic}
\usepackage{graphicx}
\expandafter\let\csname equation*\endcsname\relax
\expandafter\let\csname endequation*\endcsname\relax
\usepackage{amsmath}
\usepackage{bm}
\usepackage{textcomp}
\usepackage{multirow}
\usepackage{lscape}
\usepackage{booktabs}
\usepackage{bbm}
\usepackage[utf8]{inputenc}
\usepackage{times}
\usepackage{soul}
\usepackage{url}
\usepackage{multicol}
\usepackage[ruled,vlined]{algorithm2e}
\usepackage{lscape}

\usepackage{wrapfig}
\usepackage{colortbl}

\definecolor{gray}{rgb}{0.5,0.5,0.5}
\definecolor{darkergreen}{RGB}{21, 152, 56}

\newcommand{\norm}[1]{\left\lVert#1\right\rVert}
\newcommand{\red}[1]{\textcolor{red}{#1}}

\newcommand{\fn}[1]{\footnotesize{#1}}

\begin{document}

\title[]{Bridging the Gap Between Patient-specific and Patient-independent Seizure Prediction via Knowledge Distillation}

\author{Di Wu, Jie Yang, and Mohamad Sawan}

\address{Cutting-Edge Net of Biomedical Research and INnovation (CenBRAIN) Laboratory, School of Engineering, Westlake University, Hangzhou, 310024, China}

\address{Institute of Advanced Technology, Westlake Institute for Advanced Study, Hangzhou, 310024, China}

\ead{\mailto{yangjie@westlake.edu.cn}, \mailto{Sawan@westlake.edu.cn}}

\vspace{10pt}
\begin{abstract}
\it{Objective}. Deep neural networks (DNN) have shown unprecedented success in various brain-machine interface (BMI) applications such as epileptic seizure prediction. However, existing approaches typically train models in a patient-specific fashion due to the highly personalized characteristics of epileptic signals. Therefore, only a limited number of labeled recordings from each subject can be used for training. As a consequence, current DNN based methods demonstrate poor generalization ability to some extent due to the insufficiency of training data. On the other hand, patient-independent models attempt to utilize more patient data to train a universal model for all patients by pooling patient data together. Despite different techniques applied, results show that patient-independent models perform worse than patient-specific models due to high individual variation across patients. A substantial gap thus exists between patient-specific and patient-independent models. In this paper, we propose a novel training scheme based on knowledge distillation which makes use of a large amount of data from multiple subjects. It first distills informative features from signals of all available subjects with a pre-trained general model. A patient-specific model can then be obtained with the help of distilled knowledge and additional personalized data. \it{Significance}. The proposed training scheme significantly improves the performance of patient-specific seizure predictors and bridges the gap between patient-specific and patient-independent predictors. Four state-of-the-art seizure prediction methods are trained on the CHB-MIT sEEG database with our proposed scheme. The resulting accuracy, sensitivity, and false prediction rate show that our proposed training scheme consistently improves the prediction performance of state-of-the-art methods by a large margin.
\end{abstract}

\vspace{2pc}

\section{Introduction}
\label{sec:introduction}
Advancement in neuroscience and deep neural networks (DNN) have significantly improved the state of the arts for diagnosis and early intervention of neurological disorders such as Parkinson's, Alzheimer and epilepsy. Affecting more than 50 million people worldwide~\cite{WHO2006}, epilepsy is one of the most urgent among all neurological disorder diseases to be tackled. An epileptic seizure is characterized by a sudden, uncontrolled electrical disturbance in the brain, causing changes in behavior, movements, feelings, and sometimes loss of consciousness. To benefit the life quality of patients suffering from epilepsy seizure, numerous algorithms and systems\cite{Yang2020} have been developed to detect\cite{6634266} or forecast\cite{BouAssi2018} the incidence of seizures in high-risk situations where medical assistance is not available. Currently, state-of-the-art seizure predictions are achieved by identifying the short period before seizure occurrence based on the machine learning empowered analysis of electroencephalography signals. However, based on the prior knowledge that seizure-generating mechanisms are highly heterogeneous across patients~\cite{shorvon2011causes}, most existing machine learning models are trained in a patient-specific manner where one classifier is trained per subject \cite{Xu2020,Zhang2021} instead of a universal model for all patients. The seizure onset bio-markers may also vary for individual patients. \cite{freestone2017forward}.

\begin{figure}[t]
\centerline{\includegraphics[width=\columnwidth]{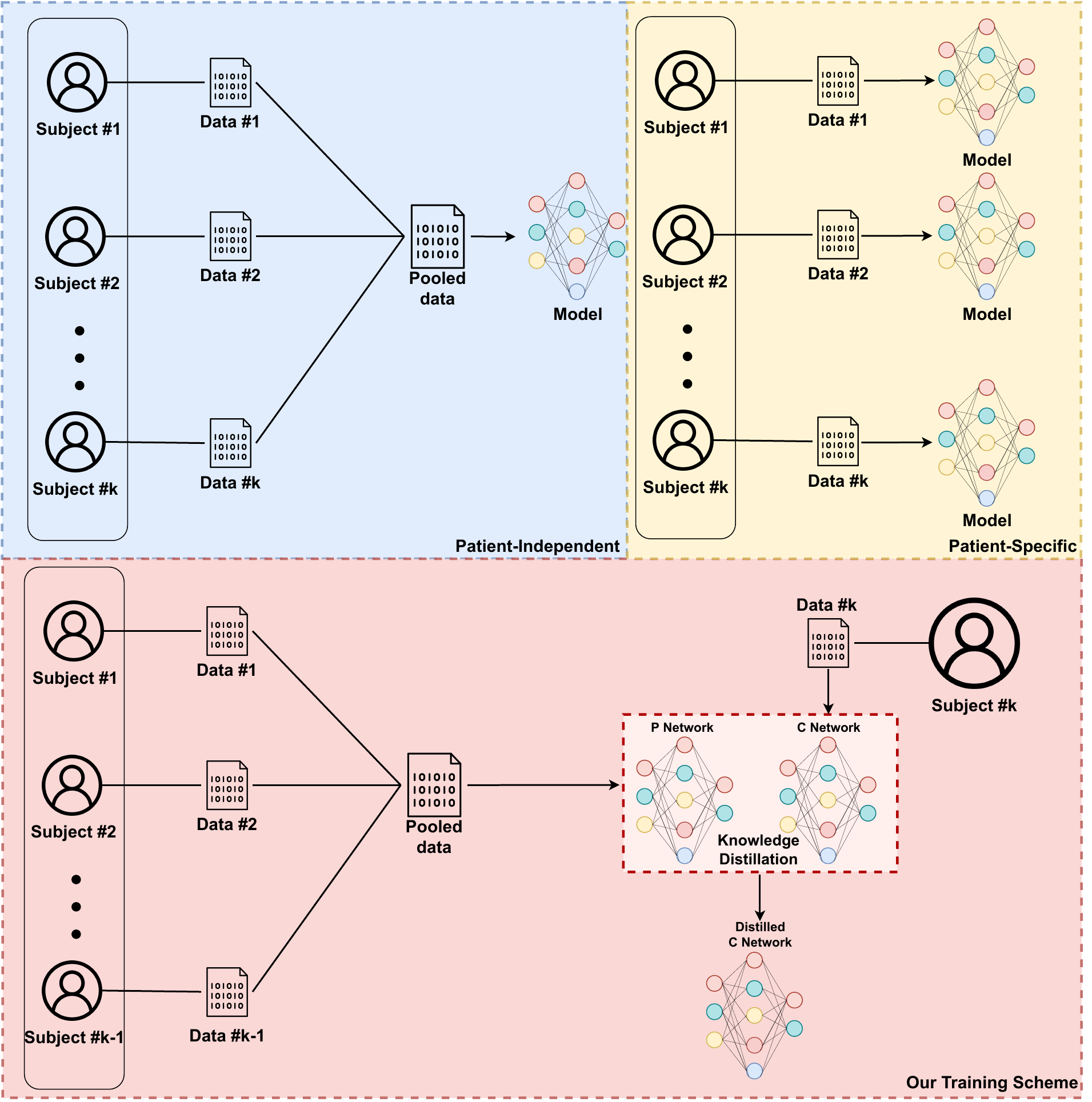}}
\caption{Comparison among patient-independent, patient-specific, and our training scheme. Patient-independent training scheme (blue region) trains a model using data pooled from all subjects. Patient-specific training scheme (yellow region) trains a unique model for each subject. Our training scheme (red region) first trains a P network using data pooled from $k-1$ subjects in the first training stage. Then during the second training stage, we randomly initialize a C network with the same structure as the P network. By applying knowledge distillation to the P network and C network using the $k^{th}$ subject's data, we generate a distilled C network that enables better prediction performance.}
\label{fig_intro}
\end{figure}
Epilepsy is a disease where the time span of actual clinically relevant behavioral symptoms is meager compared to the time span of subjects' daily life. Approximately less than 0.05\% of day-to-day time patients are suffering from actual seizure onset \cite{freestone2017forward}. Due to the low occurrence frequency nature of seizure onset and patient noncompliance, patient-specific epilepsy data might not fully cover all seizure-generating mechanisms for the individual patient during a limited recording period. Consequently, patient-specific models are usually overfitted to the limited training data (observed seizure-generating mechanisms) with poor generalization ability to new data (unseen seizure-generating mechanisms). This explains the poor prediction performance when applying patient-specific models on some individual patients.

Since manual labeling of sufficient training data for epilepsy seizures is often prohibitive, for a patient short of labeled data, there is strong motivation to build effective classifiers that can leverage rich labeled data from other patients. Some research endeavors have been made to utilize more patient data to study patient-independent models using pooled patient data for either epilepsy detection \cite{turner2014deep}\cite{9441235}\cite{YANG2020103671}\cite{7527433} or prediction \cite{9347465}. Despite different efforts of making patient-independent models possible, Roy \textit{et al.~\cite{roy2019deep} pointed out that patient-specific models often outperform patient-independent models by investigating works that incorporate both training schemes.} The problem arises that how could we solve the dilemma where using patient-specific data is insufficient for modeling complex seizure-generating mechanisms variation while using pooled patient data suffers from bio-marker inconsistency across patients.

To answer the proposed question and to bridge the gap between the patient-specific and patient-independent training scheme, we propose a novel end-to-end knowledge distillation based dual-stage training scheme which boosts the performance of individual subjects by learning seizure-generating mechanisms from other subjects. A comparison of our training scheme w.r.t patient-specific and patient-independent training scheme is provided in Figure \ref{fig_intro}. Specifically, in the first stage, we pre-train a pool network (P network) using other subjects' data to model all possible seizure-generating mechanisms regardless of the notion of patients. In the second stage, we introduce a customized network (C network) with an initial randomized condition, then both the P network and C network are iteratively optimized under a consistency constraint. The C network with knowledge distilled from the P network is used as the final classifier. Notice that in this work, we do not propose any specific neural network algorithm for seizure prediction. Instead, we provide a generic plug-and-play training scheme that could improve the performance of existing patient-specific based state-of-the-art methods. Moreover, besides the application of seizure prediction using scalp EEG (sEEG), our proposed training scheme can be applied to other neurological disorders application scenarios in the forms of electrophysiological signals such as electrocardiography (ECG) and electromyography (EMG).

The main contributions of this paper are :

\begin{itemize}
\item We bridge the gap between patient-specific and patient-independent training schemes for neurological disorders applications with a solution that remedies the inadequacy of disorder-generating mechanisms modeling and bio-marker inconsistency across patients at the same time.
\item We propose a novel end-to-end dual-stage training scheme that maximally utilizes all patients' data to enhance seizure prediction performance via knowledge distillation.   
\item We demonstrate the effectiveness of our proposed training scheme by applying it to four state-of-the-art baselines for seizure prediction on a popular open-source dataset. Experimental results show that our proposed framework constantly improves the prediction performance on all methods.
\end{itemize}

\section{Related Work}
\label{sec:related_work}
 In this section, we perform a literature survey on related works from the perspectives of transfer learning, knowledge distillation as well as seizure prediction.
\subsection{Transfer Learning}
There has been a surge of success in supervised deep learning using millions of annotations~\cite{krizhevsky2012imagenet, vaswani2017attention}. Nevertheless, such sufficient high-quality annotations are not always available in real-world applications. Thus, in such cases, knowledge transfer or transfer learning domain with abundant annotated instances to domain with limited labeled samples would be desirable. Define $\mathcal{D}$ to be a domain which consists of a feature space $\mathcal{X}$ and a marginal probability distribution $P(X)$, where $X$ is a set of data samples $X=[x_i]_{i=1}^{n}$. Transfer learning aims at improving the performance of target learners on target domains $\mathcal{D}^T$ by transferring the knowledge contained in different but related source domains $\mathcal{D}^S$. This enables us to reduce the dependence on a large number of target domain data to construct target learners. Transfer learning can be divided into three categories, i.e., transductive, inductive, and unsupervised transfer learning \cite{pan2009survey}. Roughly speaking, transductive transfer refers to the situation where the label information is not available in the target domain while label information is available in both source and target domains for inductive transfer learning. On the contrary, unsupervised transfer learning refers to the situation where labels are missing in both source and target domains. One line of researches\cite{huang2006correcting, sun2011two, 10.1145/1273496.1273521} focus on re-weighting the instances from the source domain. These re-weighted instances from the source domain are then merged with the labeled target-domain instances to train the target model. In particular, Huang \textit{et al.}~\cite{huang2006correcting} re-weights samples to match the means between the source domain and the target-domain samples in a Reproducing Kernel Hilbert Space (RKHS). Dai\textit{et al.}~\cite{10.1145/1273496.1273521} proposed to adjust the weights iteratively in the training process. Some other research endeavors have been made to transfer knowledge by measuring the distribution difference between the source domain and the target domain~\cite{yan2017mind, zellinger2017central, shen2018wasserstein}.
Maximum Mean Discrepancy (MMD) is widely adopted to measure the marginal and conditional distribution differences. Yan \textit{et al.} proposed to use a weighted version of MMD to handle class imbalance. Besides MMD, Wasserstein distance~\cite{shen2018wasserstein} is adopted for such measurement. Since our intuition is to find the correspondence of electrophysiological disorders across patients and use the knowledge learned from existing subjects as reference upon modeling for a new subject, we consider adopting knowledge distillation to address the transferring task.



\subsection{Knowledge Distillation}

Recent success in artificial intelligence, including a variety of applications in computer vision \cite{krizhevsky2012imagenet}, and natural language processing \cite{vaswani2017attention}, could be primarily contributed to the use of large-scale deep neural networks. However, for wearable or embedded systems, the deployment of such deep models is a huge challenge due to the miniature and power constraints. Knowledge distillation (KD) \cite{hinton2015distilling} was initially brought up to relieve the burden of high computational complexity and enormous memory footprint imposed by the use of heavy models. Specifically, knowledge is distilled from a large teacher model into a small student model under supervision. The intuition is that the student model learns from the teacher model to obtain a comparable or even superior performance. Besides adopting different model architecture for teacher and student model, some adopt identical architecture for both networks~\cite{yang2019snapshot,zhang2018deep, tarvainen2017mean, phuong2019distillation}. Interestingly, Hou \textit{et al.} \cite{hou2019learning} proposed to provide supervision signal from deeper layer attention maps to shallow layers of the same network. Based on different types of supervision, we could roughly divide KD methods into feature-based supervision \cite{zagoruyko2016paying,lee2018self, heo2019knowledge} and logit-based supervision \cite{islam2021breathtrack,meng2019conditional,zhang2018deep}. For logit-based methods, given a logit vector $z$ output by the last linear layer of a neural network, we formulate the logit-based distillation supervision as:
\begin{equation}
\mathfrak{L}_{kd} = \mathfrak{D}(z_t, z_s),
\end{equation}

where $\mathfrak{D}$ stands for any function that measures the divergence between logit output of teacher network $z_t$ and logit output of student network $z_s$ and $\mathfrak{L}_{kd}$ stands for knowledge distillation supervision loss. To avoid student model learning erroneous knowledge from teacher model, Meng \textit{et al.}~\cite{meng2019conditional} proposed a conditional learning scheme to allow student model to select whether to learn from teacher model or ground truth label based on the performance of the teacher model. As logits are the output of the last layer of the network, the student model lacks supervision of how intermediate processes are acquired by directly learning from the result. For feature-based methods, given an input $x$, we formulate the distillation supervision as:
\begin{equation}
\mathfrak{L}_{kd} = \mathfrak{S}(\Phi(f^{i}_{s}(x)), \Theta(f^{j}_{t}(x))),
\end{equation}
where $f^{i}_{s}(x)$ and $f^{j}_{s}(x)$ stands for the intermediate feature map of the $i^{th}$ and $j^{th}$ layer of student and teacher model respectively and $\Phi, \Theta$ are transformation functions such as a cascade of fully connected layers to match the dimension of $f^{i}_{s}(x)$ and $f^{j}_{s}(x)$, $\mathfrak{S}$ stands for any function that measures the similarity. Zagoruyko \textit{et al.}~\cite{zagoruyko2016paying} proposed to force the student model to mimic the attention maps of powerful teacher-student networks to improve the performance of CNN networks. Lee \textit{et al.}~\cite{lee2018self} proposed a more efficient distillation method by applying singular value decomposition (SVD) and radial basis function (RBF) to compress knowledge from the teacher network.

In this paper, inspired by knowledge distillation, we propose a bi-directional knowledge distillation method that distills knowledge between a pre-trained pool model and a customized model to make full use of all subjects' neurological recordings.
\subsection{Seizure detection/prediction}

Neurological disorders pose a heavy global burden and public health challenge. According to World Health Organization (WHO) \cite{WHO2006}, there are 24.3 million people who are suffering from dementia, also known as Alzheimer's Disease, with 4.6 million new cases annually. Patients with dementia gradually become highly dependent on family members and caregivers. Another widespread neurological disorder, epilepsy afflicts around 50 million individuals of which approximately up to one-third are drug-refractory \cite{camfield1996antiepileptic}. Accurate neurological disorders prediction could provide the possibility of a better life for patients.

To this end, many researches \cite{Xu2020} \cite{Zhang2021} \cite{Lawhern2018} were conducted to detect and some to predict seizures utilizing either invasive or scalp EEG. Conventional machine learning approaches are mainly support vector machine (SVM) \cite{Parvez2015} \cite{Yang2018EpilepticSP} and multi-layer perceptrons (MLP) \cite{BEHNAM2016} based. However, early methods require prior knowledge to extract hand-crafted features as input \cite{Mirowski2008}. More recently, deep learning (DL) based approaches \cite{Truong2018} \cite{TSIOURIS201824} \cite{Zhao2021} are proposed to mitigate the drawback of the manual feature extraction process being time-consuming and the features extracted lacking generalization ability. As two extensively used methods, convolutional neural networks (CNN) are used for automatic feature extracting, while recurrent neural networks (RNN) are used for better sequential relationship modeling.

Whether the method is deep learning based or whether the method takes raw data or features as input, most existing methods are trained in a patient-specific scheme. Recently, some researchers have switched attention to building subject-independent models. Zhu \textit{et al.} \cite{zhu2021unsupervised} proposed a subject-independent model for seizure prediction via adversarial training. They map the features from multiple subjects to a single subject-invariant space. They yield a result of 9.4 \% improvement in accuracy under a 1-shot classification setting on an in-house iEEG dataset with nine patients in total. Dissanayake \textit{et al.} \cite{dissanayake2021deep} proposed two deep learning structures that could learn a general function of data from multiple subjects. Specifically, there proposed models yield 88.81 \% and 91.54 \% prediction accuracy respectively on Children's Hospital of Boston-MIT (CHB-MIT) sEEG database \cite{Shoeb2010}. Zhang \textit{et al.} \cite{8994148} developed a neural network model to model purely patient-specific features using adversarial training for patient-independent seizure detection. Moreover, they came up with an attention mechanism to model the importance of each EEG channel. They yield an accuracy of 80.5 \% under the leave-one-out patient-independent setting over the Temple University Hospital EEG (TUH EEG) database \cite{obeid2016temple}.

However, most existing patient-independent researches suffers from inter-patient variation. In this paper, we provide a general training scheme that could be applied to various deep learning based methods for better seizure prediction via knowledge distillation, making use of all subjects' data.
\vspace{10pt}
\section{Methodology}
\label{sec:method}
\begin{figure*}[t]
\centerline{\includegraphics[width=\columnwidth]{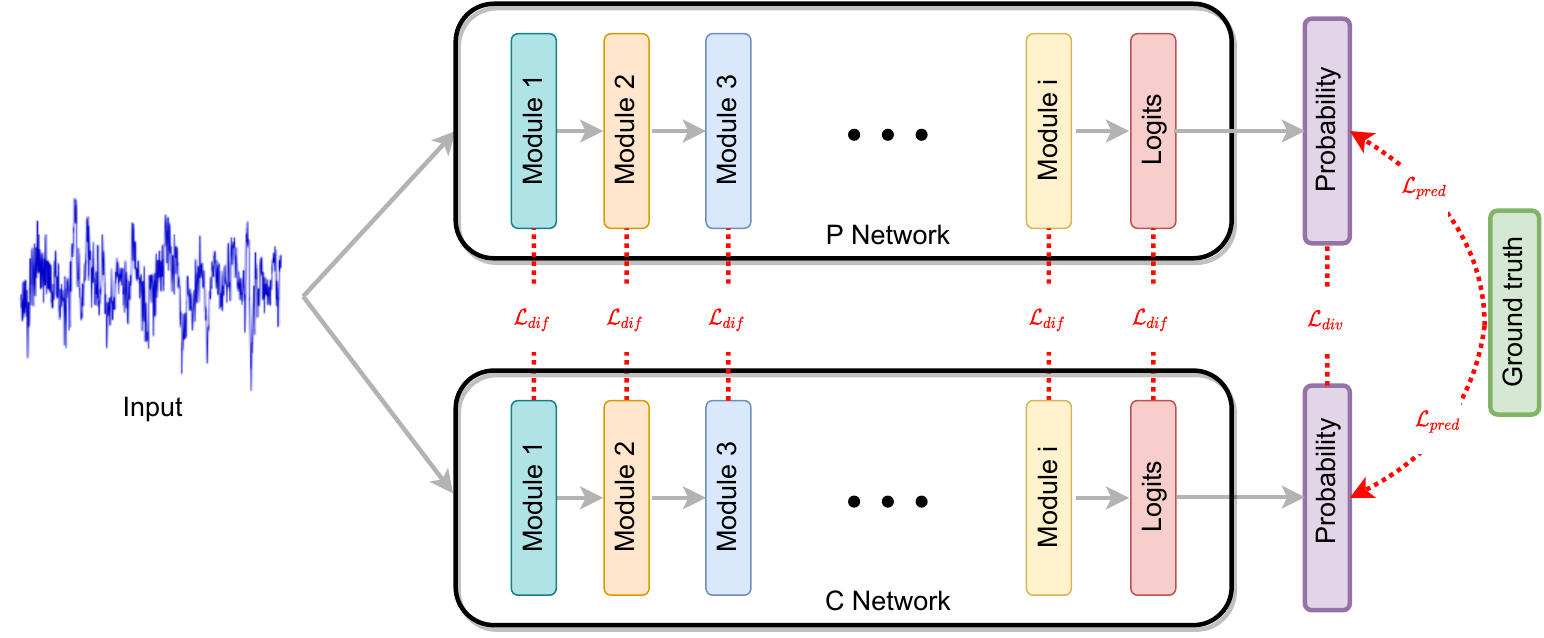}}
	\caption{
	Conceptual illustration of the second training stage of our proposed training scheme. A same training sample is fed into both P and C network. We measure $\mathcal{L}_{dif}$ across corresponding modules and $\mathcal{L}_{div}$ of the output probability distribution over P network and C network. It is important to note that neural network architecture varies greatly between works, especially in the domain of neurophysiological applications. We use the term 'module' as an abstraction of all possible layers of a neural network which could take the form of convolution, normalization, or a combination of layers. Logits represent the output of the last layer of the network.} $\mathcal{L}_{pred}$ is calculated against true label individually for both network.
	\label{fig:system}
\end{figure*}
\subsection{Preliminaries}
Our primary purpose is to fully utilize other patients' data together with one specific patient's data for better prediction performance. One important assumption is that even though subject variation exists across subjects, with enough patients data collected, similar disorder-generating mechanisms would appear between subjects. Intuitively, by modeling all seen patients' disorder-generating mechanisms, we would be able to learn from a 'patient-like-me' when a new subject's data arrives. One naive approach would be to first pre-train a model with other patients' data and then fine-tune the pre-trained model with the specific patient's data. However, fine-tuning suffers from the catastrophic forgetting, meaning that knowledge from other patients will be forgotten by the neural network rapidly upon the introduction of new data. This is because interference exists between overlapped distributed representations learned by the network~\cite{french1991using} on different data. In our framework, we first pre-train a model, which we refer to as the pool model using all other patients' data in the first stage. In the second stage, we initialize another model with the same network structure as the pool model but with random weights. The reason for adopting the same network structure for both C and P networks is that most neurophysiological signal related researches~\cite{Zhang2021, Xu2020, Lawhern2018, Truong2018} utilize different neural network architectures with huge differences. Moreover, their novelty is very much embedded in the network architecture design. Consequently, there is no golden-standard principle for shrinking different network architectures to a smaller network without jeopardizing much performance. Thus, we use identical architecture for both P and C networks in this work. We then optimize both models by asking them to learn from each other in a bi-directional fashion.
Suppose that a neurological disorder has $M$ different states and we collected data from $K$ subjects. Let $\mathcal{X}^k=\{\boldsymbol{x}^k_{1}, \boldsymbol{x}^k_{2}, ..., \boldsymbol{x}^k_{N_k}\}$ be the set of samples of the form of either electrophysiological recordings or neuroimaging images from subject $k$, where $N_k$ is number of total training samples. Let $\mathcal{Y}^k=\{\boldsymbol{y}^k_{1}, \boldsymbol{y}^k_{2}, ..., \boldsymbol{y}^k_{N_k}\}$ be the set of corresponding labels with $y_{i}$ taking value from the set $ \{1, 2, ..., M \}$. Given a neural network $f$ parameterized by $\Theta$, we denote $z^k_i = f_\Theta(x^k_i)$ as logits (the output of the last layer of a classification neural network) of sample $x^k_i$. For neurological disorder prediction, we then map $z^k_i$ to a probabilistic distribution $p^k_i$ using softmax function with temperature:
\begin{equation}
[p^k_i]_m = \frac{exp([\textit{fc}(z^k_i)]_m/T)}{\sum_mexp([\textit{fc}(z^k_i)]_m/T)},
\label{eq3}
\end{equation}
where \textit{fc} is fully connected layer which maps $z^k_i$ to $\mathbb{R}^M$, $[\cdot]_m$ means the $m^{th}$ entry and $T$ is the temperature parameter~\cite{hinton2015distilling} to control the 'softness' of the probabilistic distribution. A higher temperature produces a softer distribution while a lower temperature gives a harder distribution. We denote the predicted probability of $x^k_i$ being class $m$ as the $[p^k_i]_m$. 
Then cross entropy loss is a common choice to train the prediction task w.r.t parameters $\Theta$:
\begin{equation}
\mathcal{L}_{pred} = - \sum_{i=1}^{i=K} \sum_{j=1}^{j=N_k}\sum_{m=1}^{m=M} \mathbbm{1}({y}^i_{j}=m)log([p^k_i]_m),
\label{eq4}
\end{equation}
where the indicator function $\mathbbm{1}(\cdot)$ takes value one if the equality holds and otherwise zero.

\subsection{Training Scheme Formulation}
Next, we formalize our proposed training framework, as illustrated in Figure \ref{fig:system}. In the first training stage, we pre-train a pool model $f_{pool}$ parameterized by $\Theta_{pool}$ using all $K$ subjects data with the loss function $\mathcal{L}_{pred}$. During the second stage, given a new subject $k+1$, we aim to train a customized model $f_{cus}$ parameterized by $\Theta_{cus}$. To better model seizure-generating mechanisms and to improve prediction performance, we ask both $f_{pool}$ and $f_{cus}$ to 'learn' from each other.
Specifically, $f_{cus}$ is of the exact same architecture as $f_{pool}$ but initialized with randomized weights while $f_{pool}$ is initialized with $\Theta_{pool}$. The second training phase utilize data $\mathcal{X}^k$ from patient $k+1$ only. Given a data sample $\boldsymbol{x}^{k+1}_{i}$, denote the probabilistic distribution output by $f_{pool}$ and $f_{cus}$ as $p_{p}$ and $p_{c}$. We measure the probabilistic difference between $p_{p}$ and $p_{c}$ with Bregman divergence:
\begin{equation}
\begin{aligned}
\mathcal{L}_{div}(p_{p}, p_{c} | F(\cdot)) = \sum_{i=1}^{i=N_{k+1}} F(p_{p}({x}^{k+1}_{i})) - F(p_{c}({x}^{k+1}_{i}))\\ 
-\!\langle\nabla F(p_{c}({x}^{k+1}_{i})), p_{p}({x}^{k+1}_{i}))\!-\!F(p_{c}({x}^{k+1}_{i}))\rangle,
\end{aligned}
\label{eq5}
\end{equation}
where $F(\cdot)$ is a continuous, differentiable and convex function  defined on a closed convex set. With different choice of $F(\cdot)$, ${L}_{div}$ takes various forms. We give a few examples. With $F(p) = p^2$. In this case, $\mathcal\mathcal{L}_{div}$ is the commonly known mean squared error loss which is symmetric for $p_c$ and $p_p$.:
\begin{equation}
\begin{aligned}
\mathcal{L}_{mse}(p_{p}, p_{c}) = \norm{p_{p} - p_{c}}_2.
\label{eq6}
\end{aligned}
\end{equation}

\begin{algorithm}[t]
    \caption{The proposed training scheme}
    \SetAlgoLined
    \textbf{ Input }: 
    {\footnotesize 
    Training data set : $\mathcal{X}^k$ and $\boldsymbol{x}^{k+1}$, 
    training label set :  $\mathcal{Y}^k$ and $\boldsymbol{y}^{k+1}$,
    update flag : $flag$,
    momentum : $\phi$,
    training data set size : $N_{k+1}$,
    learning rate : $l_r$, 
    first-stage epochs : $Epoch_1$,
    second-stage epochs : $Epoch_2$,
    batch size : $B_s$,
    network structure : $N_S$,
    temperature hyper-parameter : $T$\\
    \textbf{ Output }: 
    Prediction Model : $f_\Theta$, \\
    {\# Initialization}\\ 
    Initialize pool model $f_{\Theta_{pool}}$ with $N_S$ random weight $\Theta_{pool}$\\
    Initialize cus model $f_{\Theta_{cus}}$ with $N_S$ random weight $\Theta_{cus}$\\
    Initialize update flag $flag$ to either $\textit{pool}$ or $\textit{cus}$\\
    Pre-train $f_{\Theta_{pool}}$  with $\mathcal{X}^k$ and $\mathcal{Y}^k$ using Eq.~(\ref{eq4}) for $Epoch_1$ epochs\\
    {\# epcoh loop}\\ 
    \While{$i=0$; $i<Epoch_2$; $i$++}{
        {\# batch loop}\\ 
        \While{$b=0$; $b<[ N_{k+1} /B_s]$; $b$++}{
        \uIf{$flag==pool$}{

        $\Theta_{pool} \longleftarrow \Theta_{pool} + l_r \frac{ \partial \mathcal{L}_{pool} }{\partial \Theta_{pool}}$\\
        $\Theta_{cus} \longleftarrow \phi*\Theta_{cus} + (1-\phi) * l_r \frac{ \partial \mathcal{L}_{cus} }{\partial \Theta_{cus}}$\\
        flag = cus
        }
        \Else{
        $\Theta_{cus} \longleftarrow \Theta_{cus} + l_r \frac{ \partial \mathcal{L}_{cus} }{\partial \Theta_{cus}}$\\
        $\Theta_{pool} \longleftarrow \phi*\Theta_{pool} + (1-\phi) * l_r \frac{ \partial \mathcal{L}_{pool} }{\partial \Theta_{pool}}$\\
        flag = pool
    }
        }
        }
    Return the hyper-parameter that yields better performance.
        }
\label{alg:1}
\end{algorithm}

 With $F(p) = plog(p) + (1-p)log(1-p)$, the divergence loss is then defined as an asymmetrical measurement called logistic loss:
\begin{equation}
\mathcal{L}_{div}(p_{p}, p_{c}) = p_{p}log(\frac{p_{p}}{p_{c}}) + (1-p_{p})log(\frac{(1-p_{p})}{(1-p_{c})})
\label{eq7}
\end{equation}
Besides probabilistic level measurement, we also measure the difference of intermediate features generated by the neural network as:
\begin{equation}
\mathcal{L}_{dif} = \sum_i \alpha_i\norm{z^i_{p} - z^i_{c}}_2,
\label{eq8}
\end{equation}
where $z^i_{p}$ and $z^i_{c}$ are the feature map at layer $i$ output by the two networks respectively, and $\alpha_i$ is the weight for the difference of each layer. Since the two networks are of the same structure, no additional operations are needed to match the dimension of feature maps.

The joint learning objective function for the pool model is then defined as:
\begin{equation}
\mathcal{L}_{pool} = \lambda*\mathcal{L}_{pred(pool)} + \mu*\mathcal{L}_{dif} + \nu*\mathcal{L}_{div}(p_{p}, p_{c}).
\label{eq9}
\end{equation}

Similarly, we define the learning objective function for the customized model as:
\begin{equation}
\mathcal{L}_{cus} = \lambda*\mathcal{L}_{pred(cus)} + \mu*\mathcal{L}_{dif} + \nu*\mathcal{L}_{div}(p_{c}, p_{p}).
\label{eq10}
\end{equation}

$\lambda$, $\mu$, and $\nu$ are weighting parameters to balance different loss terms. In our experiments, we set all three weighting parameters to one for simplicity since we empirically observe that the three losses naturally fall into the same level of magnitude. Notice that in the second training stage, we only use data from the ${k+1}^{th}$ subject. Conventional fine-tuning suffers from catastrophic forgetting, and thus, we disentangle the knowledge into two models with different parameters but identical structures. We update the parameters of the two models iteratively to prevent the two models from becoming identical too fast. Moreover, we propose to use momentum update~\cite{siam1992ema, 2020mocov2, nips2020byol} to address the issue of catastrophic forgetting. Formally, we denote the parameters of the network $f$ as $\Theta$ and the calculated gradient as $\nabla f$. A momentum update on $f$ is defined as:
\begin{equation}
\Theta \longleftarrow \phi*\Theta + (1-\phi)*\nabla f,
\label{eq11}
\end{equation}
where $\phi \in [0, 1)$ is a momentum coefficient. In each training iteration step, we update the parameters of one of the two models using gradient calculated from a mini-batch and update the other with momentum update and do the opposite in the next iteration. We summarize our training scheme in Algorithm \ref{alg:1}.

\section{Experiments}
\label{sec:results}
In this section, we evaluate the proposed training scheme over four state-of-the-art seizure prediction methods. We first present the experimental setup, hyper-parameters used, and training details. Then we report accuracy, sensitivity, and false positive rate (FPR) on the performance of the state-of-the-art methods trained with and without our training scheme.
\begin{figure}[t]
\centerline{\includegraphics[width=\columnwidth]{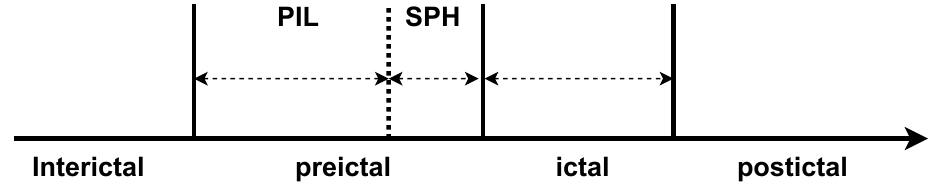}}
\caption{Definition demonstration of seizure prediction horizon (SPH) and preictal interval length (PIL). SPH is small time window between the preictal state used to generate seizure warning and the actual ictal state.}
\label{fig_seizure_states}
\end{figure}

\begin{table}[ht]
\centering
\caption{A detailed description of patients' information of CHB-MIT sEEG database. F stands for female and M stands for male. Based on the pre-processing criterion described in Section \ref{sec:data}, total training sample count is given for selected patients with others indicated by '-'.}
\label{tab:patient}

\begin{tabular}{cccccc}
\hline
Patient ID & Gender & Age & Lead seizure & Total seizure & Sample Count \\ \hline
chb01      & F      & 11                       & 3            & 7             & 354          \\
chb02      & M      & 11                       & 1            & 3             & -            \\
chb03      & F      & 14                       & 2            & 7             & -            \\
chb04      & M      & 22                       & 3            & 4             & -            \\
chb05      & F      & 7                        & 3            & 5             & 353          \\
chb06      & F      & 1.5                      & 6            & 10            & 685          \\
chb07      & F      & 14.5                     & 2            & 3             & 238          \\
chb08      & M      & 3.5                      & 3            & 5             & 355          \\
chb09      & F      & 10                       & 3            & 4             & 357          \\
chb10      & M      & 3                        & 6            & 7             & 547          \\
chb11      & F      & 12                       & 3            & 3             & -            \\
chb12      & F      & 2                        & 3            & 40            & -            \\
chb13      & F      & 3                        & 3            & 12            & -            \\
chb14      & F      & 9                        & 4            & 8             & 456          \\
chb15      & M      & 16                       & 7            & 20            & -            \\
chb16      & F      & 7                        & 1            & 10            & -            \\
chb17      & F      & 12                       & 2            & 3             & -            \\
chb18      & F      & 18                       & 3            & 6             & 247          \\
chb19      & F      & 19                       & 3            & 3             & 238          \\
chb20      & F      & 6                        & 1            & 8             & -            \\
chb21      & F      & 13                       & 1            & 4             & -            \\
chb22      & F      & 9                        & 3            & 3             & 301          \\
chb23      & F      & 6                        & 1            & 7             & -            \\ \hline
\end{tabular}%
\end{table}
\subsection{Dataset and pre-processing}
\label{sec:data}
As a proof of concept, we demonstrate the effectiveness of our proposed training scheme on the application of epilepsy seizure prediction using the publicly accessible CHB-MIT sEEG database, which contains sEEG signals from 23 epileptic patients. The database contains 637 recordings recorded from 22 bi-polar electrodes placed according to the international 10-20 system from 17 females, five males, and one person with unknown gender information. The sEEG signals were collected using the Neurofile NT digital video EEG system at a sampling rate of 256 Hz with 16-bit resolution. For each subject, seizure start time and end time are annotated by clinical experts through visual inspection. Some cases have electrodes added or removed during the EEG acquisition process. We remove these cases' data to avoid feature inconsistency caused by the variable electrodes setting during acquisition. A detailed description of all 23 cases is listed in Table \ref{tab:patient}. 

The performance analysis of an epilepsy seizure prediction system depends on the definition of the following parameters: lead seizure, seizure prediction horizon (SPH), and preictal interval length (PIL). The lead seizure is defined as a seizure preceded by a seizure-free period of length $T$. As shown in Fig. \ref{fig_seizure_states}, PIL is defined as the time period used to provide a prediction warning, while SPH is the time interval between the prediction warning and the actual seizure onset. With defined PIL and SPH, we define a seizure warning as successful if the warning appears during the time window within PIL + SPH before seizure onset. From the preictal point of view, the length of SPH is a trade-off between feeling anxious caused by a long time waiting or not having enough time for proper preparation. Unfortunately, there is no common consent on the definition of lead seizure, PIL, and SPH. In some cases, these parameters are deliberately 'picked' in order to demonstrate 'improved' prediction performance \cite{CHEN202022}.

In this work, we follow the parameter setting of Wu \textit{et al.} \cite{wu2021c2spnet}. In particular, the seizure-free period of length $T$ to define lead seizure is set as 4 hours. Patients with two or more lead seizures and at least 1-hour-long preictal time in total are selected. The SPH and PIL are chosen as 5 minutes and 30 minutes, respectively. Therefore the preictal state is defined as a 5 to 35 mins time period before seizure onset, and the interictal state is defined as 30 minutes after the seizure offset and before the PIL of the next seizure. The sliding window method is adopted to generate samples from preictal and interictal segments. Interictal samples are extracted from continuous EEG recordings with a sliding window of 20 seconds with no overlapping. A 20-second-long sliding window for the preictal samples with 25\% overlapping between two consecutive window sets is adopted to alleviate the imbalance of preictal and interictal samples. Both preictal and interictal samples are normalized by subtracting mean and dividing standard deviation.

\subsection{Experimental Setup}
\label{sec:exp_setup}
We adopt four deep learning based state-of-the-art methods and compare the performance of training with and without our proposed training scheme. In particular, two of these methods take raw EEG signals as input, while the remaining two methods adopt manually extracted features of EEG signals as input to the neural network. We briefly introduce these adopted in this paper:
\begin{itemize}
\item Lightweight solution: Zhang \textit{et al.}~\cite{Zhang2021} proposed to first calculate the Pearson correlation coefficient (PCC) \cite{00000539-201805000-00050} across all EEG channels, then the PCC matrix is then fed into a CNN based neural network for seizure prediction. Their approach is lightweight in that the size of the PCC matrix is only related to the number of channels, not the length of the signal.
\item End-to-End approach: Xu \textit{et al.}~\cite{Xu2020} proposed to use 1-dimensional (1D) kernels in the shallower layers of the CNN network and 2D kernels in the deeper layers. They claim that adopting different kernels helps to better model spatial-temporal characteristics. They use raw EEG signals as input to the CNN network.
\item EEGNet: Lawhern \textit{et al.}~\cite{Lawhern2018} proposed to utilize temporal convolution, depthwise convolution, and separable convolution for optimal spatial filtering and filter-bank construction. Their approach reduces the number of trainable parameters of the CNN network compared with previous methods. They adopt original EEG signals as input to the CNN network.
\item STFT CNN: Truong \textit{et al.}~\cite{Truong2018} proposed to first apply short-time Fourier transform (STFT) to the EEG signal, and then the 2-dimensional frequency-time matrix is fed into a 3-layer CNN for classification.
\end{itemize}

We implement all four state-of-the-art methods closely following the setup described in the original paper with Pytorch. All experiments are conducted with NVIDIA GeForce GTX 2080Ti graphics processing unit (GPU). Since there exists dataset split and pre-processing variation among these methods, to ensure a fair comparison, we adopt the dataset pre-processing and split method described in Section \ref{sec:data} for all methods. Each method is trained each baseline method for 150 epochs in a patient-specific manner. We adopt the Adam optimizer and grid search learning rate $lr \in \{1e-5, 5e-5, 1e-4, 5e-4, 1e-3, 5e-3, 1e-2\}$ and batch size $bs \in \{4, 8, 16, 32\}$. The performance is given on the test set with the hyper-parameter setup that gives the highest prediction accuracy on the validation set. For evaluation of our training scheme, we adopt the leave-one-out strategy. There are 11 patients in total selected in this study. For each baseline, we train 11 pool models and customized models, respectively. For instance, the customized model is trained on patient chb10, and the pool model is trained with all other patients' data without chb10. This is to mimic the scenario where we have ten patients' data and train a model for the $11^{th}$ patient. In addition to the leave-one-out strategy, we apply five-fold cross validation on the left-out patient's data. In specific, we randomly split the data into five equal-sized folds and use four folds as the training set and the remaining one fold as the testing set. The splitting procedure is repeated five times so that each fold is used once as testing set, and we report the mean and standard deviation of all metrics. Accuracy, sensitivity, and false prediction rate (FPR) are adopted to evaluate the performance. In the first training stage, we train each pool model with a learning rate $lr=5e-4$ and batch size $bs=32$ for 150 epochs. In the second stage, we adopt the hyper-parameter setup grid searched when training patient-specific models for baseline methods. Temperature and momentum are set to 4 and 0.995, respectively. It is worth noting that network structures vary greatly across works~\cite{Zhang2021, Xu2020, Lawhern2018, Truong2018}, the weighting parameters $\alpha$ in Equation \ref{eq8} should be treated as hyper-parameters and tuned as common practice for best performance. However, our goal is to design a generic training scheme and use seizure prediction as proof of concept in this work. Thus, we choose the output of pooling layers in all four methods as the modules for the calculation of $\mathcal{L}_{dif}$ with the minimum tuning of $\alpha$ so that $\mathcal{L}_{dif}$ is of the same level of magnitude as other two loss terms. In addition, we also compare our proposed training scheme to fine-tuning with three different settings: (1) learning rate reduction~\cite{jacobs1988increased}; (2) early stopping~\cite{prechelt1998early} and (3) interleaving training data~\cite{kirkpatrick2017overcoming}. In particular, for tine-tuning with learning rate reduction and early stopping, we first train a model using pooled data of $k-1$ subjects' data the same way we train the P network. Then, for learning rate reduction, we further train the model using the $k^{th}$ subject's data on top of the trained network with a learning rate reduction rate of 0.95, meaning the learning rate is decayed by 0.05\% at each epoch. For early stopping, we again train the model using the $k^{th}$ subject's data on top of the trained network with an early sopping criterion where the prediction performance on the validation set stops improving for five consecutive epochs. For interleaving training data, we alternate training data between $k-1$ subjects' data and the $k^{th}$ subject's data for consecutive epochs. For example, if the model is trained on $k-1$ subjects' data for the present epoch and in the next epoch, the model will be trained on the $k^{th}$ subject's data and $k-1$ subjects' data for the next next epoch. We follow the same training setup as mentioned earlier in this section for all unstated training setups.
\begin{table}[t]
\caption{Performance comparison of our proposed training scheme with fine-tuning under different settings.}
\label{tab:performace_comparison}
\resizebox{\textwidth}{!}{%
\begin{tabular}{cccccc} 
\hline
\textbf{Loss}                                                                     & \textbf{Metrics} & \textbf{Xu~\cite{Xu2020}}                                      & \textbf{Zhang~\cite{Zhang2021}}                                  & \textbf{Lawhern~\cite{Lawhern2018}}               & \textbf{Truong~\cite{Truong2018}}                                   \\ 
\hline
\multirow{3}{*}{\begin{tabular}[c]{@{}c@{}}Learning rate\\reduction~\cite{jacobs1988increased}\end{tabular}} & Accuracy (\%)    & 84.8$\pm$3.6                                      & 89.9$\pm$0.8                                     & 87.3$\pm$1.3                    & 83.6$\pm$1.5                                       \\
                                                                                  & Sensitivity (\%) & 85.7$\pm$3.3                                      & 93.3$\pm$1.4                                     & 90.6$\pm$2.5                    & 90.4$\pm$3.5                                       \\
                                                                                  & FPR (/h)         & 0.18$\pm$0.08                                     & 0.13$\pm$0.03                                    & 0.16$\pm$0.04                   & 0.22$\pm$0.05                                      \\ 
\hline
\multirow{3}{*}{\begin{tabular}[c]{@{}c@{}}Early\\stopping~\cite{prechelt1998early}\end{tabular}}          & Accuracy (\%)    & 84.6$\pm$3.5                                      & 90.1$\pm$0.8                                     & 87.7$\pm$1.2                    & 83.7$\pm$1.4                                       \\
                                                                                  & Sensitivity (\%) & 85.3$\pm$3.2                                      & 93.5$\pm$1.3                                     & 90.6$\pm$2.7                    & 90.7$\pm$3.4                                       \\
                                                                                  & FPR (/h)         & 0.18$\pm$0.07                                     & 0.14$\pm$0.03                                    & 0.15$\pm$0.04                   & 0.21$\pm$0.06                                      \\ 
\hline
\multirow{3}{*}{\begin{tabular}[c]{@{}c@{}}Interleaving\\data~\cite{kirkpatrick2017overcoming}\end{tabular}}       & Accuracy (\%)    & 81.3$\pm$5.9                                      & 87.5$\pm$1.4                                     & 84.4$\pm$3.9                    & 82.1$\pm$1.8                                       \\
                                                                                  & Sensitivity (\%) & 90.4$\pm$7.1                                      & 92.9$\pm$2.2                                     & 88.2$\pm$4.1                    & 88.2$\pm$3.9                                       \\
                                                                                  & FPR (/h)         & 0.25$\pm$0.09                                     & 0.17$\pm$0.04                                    & 0.18$\pm$0.10                   & 0.24$\pm$0.07                                      \\ 
\hline
\multirow{3}{*}{\begin{tabular}[c]{@{}c@{}}This\\work\end{tabular}}               & Accuracy (\%)    & \textbf{\textbf{89.3}}$\pm$\textbf{\textbf{3.0}}  & \textbf{92.9}$\pm$\textbf{0.8}                   & \textbf{90.1}$\pm$\textbf{0.9}  & \textbf{\textbf{88.3}}$\pm$\textbf{\textbf{1.1}}   \\
                                                                                  & Sensitivity (\%) & \textbf{\textbf{92.3}}$\pm$\textbf{\textbf{2.4}}  & \textbf{\textbf{94.8}}$\pm$\textbf{\textbf{1.2}} & \textbf{93.7}$\pm$\textbf{1.7}  & \textbf{92.1}$\pm$\textbf{2.9}                     \\
                                                                                  & FPR (/h)         & \textbf{\textbf{0.14}}$\pm$\textbf{\textbf{0.06}} & \textbf{0.10}$\pm$\textbf{0.02}                  & \textbf{0.13}$\pm$\textbf{0.02} & \textbf{\textbf{0.18}}$\pm$\textbf{\textbf{0.04}}  \\
\hline
\end{tabular}%
}
\end{table}
\begin{table*}[ht]
\centering
\caption{Prediction performance comparison of baselines trained with and without our training scheme. The baseline method with a superscript $*$ indicates trained with the proposed training scheme.}
\label{tab:result}
\resizebox{\textwidth}{!}{%
\begin{tabular}{cccccccccc} 
\hline
\textbf{Patient ID}      & \textbf{Metrics } & \textbf{Xu~\cite{Xu2020}}                                            & \textbf{Xu$^*$}                 & \textbf{Zhang~\cite{Zhang2021}} & \textbf{Zhang$^*$}              & \textbf{Lawhern~\cite{Lawhern2018}} & \textbf{Lawhern$^*$}            & \textbf{Truong~\cite{Truong2018}} & \textbf{Truong$^*$}              \\ 
\hline
\multirow{3}{*}{chb01}   & Accuracy (\%)     & 98.9$\pm$0.8                                            & 99.4$\pm$0.3                    & 98.6$\pm$0.4    & 99.3$\pm$0.4                    & 98.2$\pm$0.7      & 99.4$\pm$0.4                    & 96.5$\pm$0.8     & 98.2$\pm$0.6                     \\
                         & Sensitivity (\%)  & 98.7$\pm$1.3                                            & 99.4$\pm$0.2                    & 99.7$\pm$0.5    & 99.2$\pm$0.2                    & 99.5$\pm$0.6      & 99.4$\pm$0.2                    & 98.2$\pm$0.3     & 99.4$\pm$0.2                     \\
                         & FPR (/h)          & 0.01$\pm$0.02                                           & 0.01$\pm$0.01                   & 0.03$\pm$0.01   & 0.01$\pm$0.01                   & 0.03$\pm$0.02     & 0.01$\pm$0.01                   & 0.03$\pm$0.02    & 0.02$\pm$0.01                    \\ 
\hline
\multirow{3}{*}{chb05}   & Accuracy (\%)     & 84.3$\pm$4.6                                            & 89.5$\pm$3.2                    & 86.9$\pm$2.3    & 90.5$\pm$2.1                    & 87.6$\pm1$1.1     & 91.2$\pm$1.2                    & 80.3$\pm$1.3     & 83.5$\pm$0.8                     \\
                         & Sensitivity (\%)  & 87.7$\pm$3.2                                            & 92.5$\pm$1.5                    & 92.3$\pm$0.9    & 93.3$\pm$1.6                    & 94.1$\pm$3.5      & 90.9$\pm$1.8                    & 90.5$\pm$3.5     & 93.6$\pm$2.5                     \\
                         & FPR (/h)          & 0.18$\pm$0.05                                           & 0.16$\pm$0.03                   & 0.21$\pm$0.04   & 0.18$\pm$0.04                   & 0.22$\pm$0.05     & 0.17$\pm$0.03                   & 0.35$\pm$0.04    & 0.26$\pm$0.03                    \\ 
\hline
\multirow{3}{*}{chb06}   & Accuracy (\%)     & 75.8$\pm$3.4                                            & 81.9$\pm$2.2                    & 80.1$\pm$1.0    & 83.2$\pm$1.4                    & 71.1$\pm$0.7      & 75.9$\pm$0.5                    & 71.5$\pm$1.2     & 76.2$\pm$0.6                     \\
                         & Sensitivity (\%)  & 78.0$\pm$3.8                                            & 86.3$\pm$2.5                    & 87.1$\pm$2.8    & 88.8$\pm$2.4                    & 78.2$\pm$3.8      & 82.4$\pm$2.7                    & 82.6$\pm$ 6.0    & 87.2$\pm$~5.4                    \\
                         & FPR (/h)          & 0.26$\pm$0.04                                           & 0.18$\pm$0.03                   & 0.27$\pm$0.02   & 0.24$\pm$0.02                   & 0.33$\pm$0.04     & 0.29$\pm$0.02                   & 0.40$\pm$0.05    & 0.34$\pm$0.04                    \\ 
\hline
\multirow{3}{*}{chb07}   & Accuracy (\%)     & 84.8$\pm$1.7                                            & 89.4$\pm$1.3                    & 90.6$\pm$0.7    & 93.8$\pm$0.5                    & 85.2$\pm$0.8      & 92.7$\pm$0.7                    & 81.4$\pm$2.4     & 89.3$\pm$2.6                     \\
                         & Sensitivity (\%)  & 89.8$\pm$2.8                                            & 92.6$\pm$2.4                    & 94.0$\pm$1.6    & 98.1$\pm$1.4                    & 95.7$\pm$0.2      & 94.6$\pm$0.2                    & 91.1$\pm$6.2     & 94.0$\pm$5.4                     \\
                         & FPR (/h)          & 0.20$\pm$0.04                                           & 0.15$\pm$0.03                   & 0.13$\pm$0.02   & 0.14$\pm$0.02                   & 0.25$\pm$0.02     & 0.16$\pm$0.02                   & 0.29$\pm$0.08    & 0.19$\pm$0.05                    \\ 
\hline
\multirow{3}{*}{chb08}   & Accuracy (\%)     & 91.8$\pm$1.5                                            & 92.6$\pm$1.4                    & 91.1$\pm$0.3    & 94.6$\pm$0.2                    & 87.5$\pm$0.9      & 92.4$\pm$0.5                    & 88.0$\pm$1.2     & 92.6$\pm$0.7                     \\
                         & Sensitivity (\%)  & 96.7$\pm$1.7                                            & 96.9$\pm$1.8                    & 97.4$\pm$0.5    & 98.1$\pm$0.3                    & 92.6$\pm$3.3      & 98.3$\pm$2.5                    & 96.9$\pm$0.6     & 96.9$\pm$0.6                     \\
                         & FPR (/h)          & 0.16$\pm$0.06                                           & 0.16$\pm$0.06                   & 0.19$\pm$0.09   & 0.14$\pm$0.05                   & 0.22$\pm$0.04     & 0.13$\pm$0.02                   & 0.26$\pm$0.03    & 0.24$\pm$0.03                    \\ 
\hline
\multirow{3}{*}{chb09}   & Accuracy (\%)     & 92.4$\pm$4.2                                            & 95.4$\pm$3.4                    & 95.0$\pm$0.3    & 97.1$\pm$0.4                    & 92.9$\pm$0.7      & 93.4$\pm$0.3                    & 89.9$\pm$1.2     & 93.4$\pm$1.3                     \\
                         & Sensitivity (\%)  & 93.6$\pm$3.6                                            & 95.7$\pm$2.1                    & 95.9$\pm$0.6    & 96.2$\pm$0.4                    & 96.5$\pm$2.0      & 98.4$\pm$1.7                    & 96.0$\pm$2.4     & 95.8$\pm$2.2                     \\
                         & FPR (/h)          & 0.10$\pm$0.14                                           & 0.06$\pm$0.05                   & 0.06$\pm$0.01   & 0.04$\pm$0.01                   & 0.12$\pm$0.03     & 0.07$\pm$0.02                   & 0.18$\pm$0.05    & 0.16$\pm$0.05                    \\ 
\hline
\multirow{3}{*}{chb10}   & Accuracy (\%)     & 82.5$\pm$1.6                                            & 85.2$\pm$1.4                    & 90.6$\pm$0.6    & 93.8$\pm$0.3                    & 88.8$\pm$1.3      & 91.1$\pm$0.9                    & 81.8$\pm$0.8     & 86.1$\pm$0.9                     \\
                         & Sensitivity (\%)  & 84.6$\pm$3.6                                            & 86.4$\pm$2.4                    & 94.4$\pm$0.8    & 95.9$\pm$0.4                    & 89.5$\pm$3.9      & 90.5$\pm$2.4                    & 81.7$\pm$3.4     & 87.5$\pm$2.8                     \\
                         & FPR (/h)          & 0.19$\pm$0.04                                           & 0.14$\pm$0.02                   & 0.13$\pm$0.01   & 0.11$\pm$0.01                   & 0.12$\pm$0.02     & 0.10$\pm$0.01                   & 0.18$\pm$0.03    & 0.15$\pm$0.02                    \\ 
\hline
\multirow{3}{*}{chb14}   & Accuracy (\%)     & 59.7$\pm$1.8                                            & 72.9$\pm$0.8                    & 80.7$\pm$1.2    & 85.8$\pm$1.4                    & 66.6$\pm$4.3      & 77.4$\pm$3.7                    & 61.6$\pm$3.3     & 73.4$\pm$2.4                     \\
                         & Sensitivity (\%)  & 18.2$\pm$14.6                                           & 71.2$\pm$10.2                   & 81.8$\pm$4.3    & 81.8$\pm$4.1                    & 58.4$\pm$2.1      & 79.2$\pm$2.3                    & 61.1$\pm$13.0    & 70.9$\pm$7.7                     \\
                         & FPR (/h)          & 0.12$\pm$0.12                                           & 0.14$\pm$0.09                   & 0.20$\pm$0.06   & 0.19$\pm$0.04                   & 0.26$\pm$0.10     & 0.21$\pm$0.06                   & 0.38$\pm$0.15    & 0.28$\pm$0.12                    \\ 
\hline
\multirow{3}{*}{chb18}   & Accuracy (\%)     & 96.7$\pm$0.4                                            & 97.9$\pm$0.4                    & 97.8$\pm$0.4    & 98.5$\pm$0.3                    & 96.6$\pm$0.8      & 99.1$\pm$0.4                    & 95.2$\pm$1.2     & 97.6$\pm$0.4                     \\
                         & Sensitivity (\%)  & 97.6$\pm$0.8                                            & 98.6$\pm$0.2                    & 98.0$\pm$0.4    & 99.8$\pm$0.2                    & 97.3$\pm$1.0      & 98.0$\pm$0.5                    & 96.9$\pm$1.6     & 97.1$\pm$1.5                     \\
                         & FPR (/h)          & \begin{tabular}[c]{@{}c@{}}0.04$\pm$0.00\\\end{tabular} & 0.02$\pm$0.00                   & 0.04$\pm$0.01   & 0.02$\pm$0.00                   & 0.04$\pm$0.02     & 0.01$\pm$0.01                   & 0.07$\pm$0.03    & 0.06$\pm$0.03                    \\ 
\hline
\multirow{3}{*}{chb19}   & Accuracy (\%)     & 83.3$\pm$19.2                                           & 92.5$\pm$17.1                   & 91.9$\pm$0.8    & 95.7$\pm$0.4                    & 99.0$\pm$0.0      & 100$\pm$0.0                     & 92.9$\pm$0.8     & 97.4$\pm$0.9                     \\
                         & Sensitivity (\%)  & 97.9$\pm$4.3                                            & 96.4$\pm$2.4                    & 96.1$\pm$0.6    & 99.3$\pm$0.5                    & 100$\pm$0.0       & 100$\pm$0.0                     & 94.9$\pm$2.2     & 96.8$\pm$1.4                     \\
                         & FPR (/h)          & 0.31$\pm$0.35                                           & 0.24$\pm$0.25                   & 0.12$\pm$0.02   & 0.08$\pm$0.02                   & 0.02$\pm$0.00     & 0.00$\pm$0.00                   & 0.09$\pm$0.02    & 0.04$\pm$0.02                    \\ 
\hline
\multirow{3}{*}{chb22}   & Accuracy (\%)     & 82.4$\pm$1.9                                            & 85.6$\pm$1.3                    & 86.1$\pm$1.2    & 89.1$\pm$1.0                    & 85.5$\pm$3.0      & 87.7$\pm$1.5                    & 78.5$\pm$1.1     & 83.2$\pm$0.7                     \\
                         & Sensitivity (\%)  & 99.4$\pm$0.8                                            & 99.8$\pm$0.4                    & 90.0$\pm$2.3    & 92.6$\pm$1.4                    & 93.8$\pm$5.8      & 98.6$\pm$3.9                    & 97.2$\pm$4.1     & 94.0$\pm$3.5                     \\
                         & FPR (/h)          & 0.36$\pm$0.04                                           & 0.29$\pm$0.04                   & 0.16$\pm$0.06   & 0.13$\pm$0.04                   & 0.24$\pm$0.09     & 0.31$\pm$0.07                   & 0.42$\pm$0.06    & 0.30$\pm$0.05                    \\ 
\hline
\multirow{3}{*}{Average} & Accuracy (\%)     & 84.8$\pm$3.7                                            & \textbf{89.3}$\pm$\textbf{3.0}  & 89.9$\pm$0.8    & \textbf{92.9}$\pm$\textbf{0.8}  & 87.2$\pm$1.3      & \textbf{90.1}$\pm$\textbf{0.9}  & 83.4$\pm$1.4     & \textbf{88.3}$\pm$\textbf{1.1}   \\
                         & Sensitivity (\%)  & 85.6$\pm$3.4                                            & \textbf{92.3}$\pm$\textbf{2.4}   & 93.3$\pm$1.4    & \textbf{94.8}$\pm$\textbf{1.2}  & 90.5$\pm$2.4      & \textbf{93.7}$\pm$\textbf{1.7}  & 89.7$\pm$3.9     & \textbf{92.1}$\pm$\textbf{2.9}   \\
                         & FPR (/h)          & 0.18$\pm$0.08                                           & \textbf{0.14}$\pm$\textbf{0.06} & 0.14$\pm$0.03   & \textbf{0.10}$\pm$\textbf{0.02} & 0.17$\pm$0.04     & \textbf{0.13}$\pm$\textbf{0.02} & 0.24$\pm$0.05    & \textbf{0.18}$\pm$\textbf{0.04}  \\
\hline
\end{tabular}%
}
\end{table*}
\subsection{Results}
To demonstrate the effectiveness of the proposed training scheme, we report prediction accuracy, sensitivity, and FPR over all patients listed in Table \ref{tab:patient} of all methods trained with and without the proposed training scheme in Table \ref{tab:result}. We refer to training without our proposed training scheme as the regular training scheme. For each method, the better average performance between regular and our training scheme is marked in bold. As can be seen, our training scheme consistently improves the performance of all baseline methods on all patients, illustrating the effectiveness of the proposed training scheme. Specifically, for patients with relatively poor prediction performance under a regular training scheme such as chb14, our training scheme improves the performance of End-to-End approach by 13.2\% in terms of prediction accuracy and 53\% in sensitivity. This suggests that by learning knowledge from other subjects, we can improve the prediction performance by a large margin. Since Even for patients with already relatively high prediction performance such as chb01, our training scheme still brings about improvement, although incremental. In summary, our training scheme yields an average improvement of 3.8\% and 3.5\% in accuracy and sensitivity, respectively. Moreover, our training scheme reduces the FPR by an average value of 0.05/h. We compare the performance of our proposed training scheme over fine-tuning based methods by reporting the average value of all metrics over all patients under the leave-one-out strategy in Table \ref{tab:performace_comparison}. As can be seen, fine-tuning with learning rate reduction and early stopping yield similar performance as the regular training scheme. However, applying interleaving data brings a negative impact on the prediction performance. We observe that the loss fluctuates drastically during the training of interleaving data, and we hypothesize that this is caused by the distribution difference among different patient's data. Our proposed training scheme outperforms naive fine-tuning methods by a large margin.
\begin{table}[t]
\caption{{Ablation study on the loss components. $\mathcal{L}_{pred}$ means training for seizure prediction task alone which is equivalent to the regualr training scheme. We than add our proposed $\mathcal{L}_{div}$ and $\mathcal{L}_{dif}$ one by one onto $\mathcal{L}_{pred}$.}}
\label{tab:ablation_loss2}
\resizebox{\textwidth}{!}{%
\begin{tabular}{cccccc} 
\hline
\textbf{Loss}                                                                & \textbf{Metrics} & \textbf{Xu~\cite{Xu2020}}                    & \textbf{Zhang~\cite{Zhang2021} }                 & \textbf{Lawhern~\cite{Lawhern2018} }               & \textbf{Truong~\cite{Truong2018} }                 \\ 
\hline
\multirow{3}{*}{$\mathcal{L}_{pred}$}                                        & Accuracy (\%)    & 84.8$\pm$3.7                    & 89.9$\pm$0.8                    & 87.2$\pm$1.3                    & 83.4$\pm$1.4                     \\
                                                                             & Sensitivity (\%) & 85.6$\pm$3.4                    & 93.3$\pm$1.4                    & 90.5$\pm$2.4                    & 89.7$\pm$3.9                     \\
                                                                             & FPR (/h)         & 0.18$\pm$0.08                   & 0.14$\pm$0.03                   & 0.17$\pm$0.04                   & 0.24$\pm$0.05     \\ 
\hline
\multirow{3}{*}{$\mathcal{L}_{div} + \mathcal{L}_{pred}$}                    & Accuracy (\%)    & 88.5$\pm$3.5                    & 92.1$\pm$0.8                    & 89.5$\pm$1.2                    & 87.3$\pm$1.4                     \\
                                                                             & Sensitivity (\%) & 89.7$\pm$2.9                    & 93.8$\pm$1.3                    & 92.7$\pm$2.2                    & 91.7$\pm$3.6                     \\
                                                                             & FPR (/h)         & 0.15$\pm$0.08                   & 0.13$\pm$0.03                   & 0.15$\pm$0.03                   & 0.21$\pm$3.7                     \\ 
\hline
\multirow{3}{*}{$\mathcal{L}_{dif} + \mathcal{L}_{pred}$}                    & Accuracy (\%)    & 87.4$\pm$3.2                    & 91.6$\pm$0.8                    & 89.1$\pm$1.1                    & 85.4$\pm$1.2                     \\
                                                                             & Sensitivity (\%) & 89.2$\pm$2.6                    & 93.9$\pm$1.2                    & 91.4$\pm$2.1                    & 90.5$\pm$3.4                     \\
                                                                             & FPR (/h)         & 0.16$\pm$0.07                   & 0.13$\pm$0.02                   & 0.15$\pm$0.02                   & 0.20$\pm$0.04                    \\ 
\hline
\multirow{3}{*}{$\mathcal{L}_{dif} + \mathcal{L}_{div} +\mathcal{L}_{pred}$} & Accuracy (\%)    & \textbf{89.3}$\pm$\textbf{3.0}  & \textbf{92.9}$\pm$\textbf{0.8}  & \textbf{90.1}$\pm$\textbf{0.9}  & \textbf{88.3}$\pm$\textbf{1.1}   \\
                                                                             & Sensitivity (\%) & \textbf{92.3}$\pm$\textbf{2.4}  & \textbf{94.8}$\pm$\textbf{1.2}  & \textbf{93.7}$\pm$\textbf{1.7}  & \textbf{92.1}$\pm$\textbf{2.9}   \\
                                                                             & FPR (/h)         & \textbf{0.14}$\pm$\textbf{0.06} & \textbf{0.10}$\pm$\textbf{0.02} & \textbf{0.13}$\pm$\textbf{0.02} & \textbf{0.18}$\pm$\textbf{0.04}  \\
\hline
\end{tabular}%
}
\end{table}
\begin{table}[]
\caption{Ablation study on the choice of loss function for Bregman divergence.}
\label{tab:ablation_loss}
\resizebox{\textwidth}{!}{%
\begin{tabular}{cccccc} 
\hline
\textbf{Loss}                                                            & \textbf{Metrics} & \textbf{Xu~\cite{Xu2020}}                    & \textbf{Zhang~\cite{Zhang2021}}                 & \textbf{Lawhern~\cite{Lawhern2018}} & \textbf{Truong~\cite{Truong2018}}                 \\ 
\hline
\multirow{3}{*}{MSE}                                                     & Accuracy (\%)    & 87.3$\pm$3.4                    & 91.2$\pm$0.9                    & 89.4$\pm$1.3      & 83.6$\pm$1.5                     \\
                                                                         & Sensitivity (\%) & 87.8$\pm$2.7                    & 93.8$\pm$1.2                    & 92.5$\pm$1.8      & 90.4$\pm$3.5                     \\
                                                                         & FPR (/h)         & 0.16$\pm$0.07                   & 0.13$\pm$0.02                   & 0.14$\pm$0.02     & 0.22$\pm$0.05                    \\ 
\hline
\multirow{3}{*}{\begin{tabular}[c]{@{}c@{}}KL\\ Divergence\end{tabular}} & Accuracy (\%)    & 88.1$\pm$3.3                    & \textbf{93.1}$\pm$\textbf{0.8}  & 89.4$\pm$0.9      & 87.6$\pm$1.2                     \\
                                                                         & Sensitivity (\%) & 92.3$\pm$2.4                    & 93.5$\pm$1.1                    & 93.2$\pm$1.8      & \textbf{92.7}$\pm$\textbf{2.8}   \\
                                                                         & FPR (/h)         & 0.15$\pm$0.06                   & \textbf{0.09}$\pm$\textbf{0.02} & 0.14$\pm$0.02     & 0.19$\pm$0.04                    \\ 
\hline
\multirow{3}{*}{\begin{tabular}[c]{@{}c@{}}Logistic\\ Loss\end{tabular}} & Accuracy (\%)    & \textbf{89.3}$\pm$\textbf{3.0}  & 92.9$\pm$0.8                    & 90.1$\pm$0.9      & \textbf{88.3}$\pm$\textbf{1.1}   \\
                                                                         & Sensitivity (\%) & \textbf{92.3}$\pm$\textbf{2.4}  & \textbf{94.8}$\pm$\textbf{1.2}  & 93.7$\pm$1.7      & 92.1$\pm$2.9                     \\
                                                                         & FPR (/h)         & \textbf{0.14}$\pm$\textbf{0.06} & 0.10$\pm$0.02                   & 0.13$\pm$0.02     & \textbf{0.18}$\pm$\textbf{0.04}  \\
\hline
\end{tabular}%
}
\end{table}
\newcommand{\redbf}[1]{\red{\bf{\fn{(#1)}}}}
\begin{table}[]
\caption{Ablation study on the temperature hyper-parameter. Regular approach indicates the regular training method of all baseline methods. The following three rows show the performance using our proposed training scheme with different temperature.}
\label{tab:ablation_temperature}
\resizebox{\textwidth}{!}{%
\begin{tabular}{cccccc} 
\hline
\multicolumn{1}{l}{\textbf{Temperature}}                                    & \textbf{Metrics} & \textbf{Xu~\cite{Xu2020}}  & \textbf{Zhang~\cite{Zhang2021}} & \textbf{Lawhern\cite{Lawhern2018}} & \textbf{Truong\cite{Truong2018}}                                   \\ 
\hline
\multirow{3}{*}{\begin{tabular}[c]{@{}c@{}}Regular\\ Approach\end{tabular}} & Accuracy (\%)    & 84.8$\pm$3.7  & 89.9$\pm$0.8    & 87.2$\pm$1.3      & 83.4$\pm$1.4                                       \\
                                                                            & Sensitivity (\%) & 85.6$\pm$3.4  & 93.3$\pm$1.4    & 90.5$\pm$2.4      & 89.7\textbackslash{}$pm$3.9                        \\
                                                                            & FPR (/h)         & 0.18$\pm$0.08 & 0.14$\pm$0.03   & 0.17$\pm$0.04     & 0.24$\pm$0.05                                      \\ 
\hline
\multirow{3}{*}{$T$ = 1}                                                    & Accuracy (\%)    & 86.4$\pm$3.4  & 90.4$\pm$0.9    & 87.7$\pm$1.2      & 85.1$\pm$1.5                                       \\
                                                                            & Sensitivity (\%) & 86.3$\pm$2.4  & 93.7$\pm$1.3    & 91.3$\pm$2.2      & 90.4$\pm$3.5                                       \\
                                                                            & FPR (/h)         & 0.16$\pm$0.08 & 0.13$\pm$0.03   & 0.15$\pm$0.03     & 0.22$\pm$0.05                                      \\ 
\hline
\multirow{3}{*}{$T$ = 4}                                                    & Accuracy (\%)    & 89.3$\pm$3.0  & 92.9$\pm$0.8    & 90.1$\pm$0.9      & 88.3$\pm$1.1                                       \\
                                                                            & Sensitivity (\%) & 92.3$\pm$2.4  & 94.8$\pm$1.2    & 93.7$\pm$1.7      & 92.1$\pm$2.9                                       \\
                                                                            & FPR (/h)         & 0.14$\pm$0.06 & 0.10$\pm$0.02   & 0.13$\pm$0.02     & 0.18$\pm$0.04\textcolor[rgb]{0.365,0.408,0.475}{}  \\ 
\hline
\multirow{3}{*}{$T$ = 8}                                                    & Accuracy (\%)    & 88.3$\pm$3.1  & 92.3$\pm$0.8    & 89.3$\pm$0.9      & 86.3$\pm$1.4                                       \\
                                                                            & Sensitivity (\%) & 91.6$\pm$2.5  & 93.4$\pm$1.3    & 92.1$\pm$1.7      & 90.2$\pm$4.0                                       \\
                                                                            & FPR (/h)         & 0.15$\pm$0.06 & 0.15$\pm$0.02   & 0.15$\pm$0.03     & 0.20$\pm$0.05                                      \\
\hline
\end{tabular}%
}
\end{table}
\subsection{Ablation Study}
Ablation studies are performed on the same dataset as described in Section \ref{sec:data} with same experimental setup as in described in Section \ref{sec:exp_setup}. We report the average performance of all patients under the leave-one-out strategy. To analyze the effectiveness and to provide more insights into our proposed training scheme, we provide an ablation study from the following aspects:

(W/o $\mathcal{L}_{div}$ or $\mathcal{L}_{dif}$): During the second training phase, the overall loss function to train each network is given as $\mathcal{L} = \mathcal{L}_{pred} + \mathcal{L}_{dif} + \mathcal{L}_{div}$. $\mathcal{L}_{pred}$ is a task-related loss function for certain neurological disorder prediction, which is seizure prediction in our case and thus cannot be removed. Training with $\mathcal{L}_{pred}$ alone is equivalent to training without our proposed training scheme. So we add $\mathcal{L}_{dif}$ and $\mathcal{L}_{div}$ one by one onto $\mathcal{L}_{pred}$ and show how these two losses affect the training performance. We show results in Table \ref{tab:ablation_loss2}. The best performance is marked in bold. We can see that adding each of $\mathcal{L}_{dif}$ and $\mathcal{L}_{div}$ improves the prediction performance, which indicates the effectiveness of both loss functions. Also, we see that training with both $\mathcal{L}_{dif}$ and $\mathcal{L}_{div}$ further improves the prediction performance.

The choice of loss function for $\mathcal{L}_{div}$: Next, we specify three different choices of $F(\cdot)$ for Bregman divergence. Based on the choice of $F(\cdot)$, the resulting Bregman divergence losses are mean squared error loss, logistic loss, and Kullback–Leibler (KL) divergence. The experimental result is given in Table \ref{tab:ablation_loss}. The best performance of among the results using different loss functions is marked in bold. We can observe from this table that KL divergence and the logistic loss yield similar performance while MSE yields slightly worse results than the other two losses. We hypothesize that KL divergence and logistic loss are asymmetric and thus work better with the proposed training scheme. In particular, our training scheme distills knowledge bi-directionally, we update the parameters of one of the two models using gradient calculated from a mini-batch and update the other with momentum update and do the opposite in the next iteration. The asymmetric nature of the loss better enables learning iteratively rather than the co-learning effect of the mean squared error (MSE) loss. Nevertheless, the proposed training scheme is robust to the choice of loss functions.

The temperature $T$: The pool model embeds class relationships prior into the customized model's logit layer geometry with proper tuning of the temperature hyper-parameter \cite{tang2020understanding}. We show the impact of different temperatures on our proposed training scheme. The result is shown in Table \ref{tab:ablation_temperature}. The number in the parentheses indicates performance comparison against the baseline method trained with a regular approach. Green color indicates performance improvement while red color indicates performance degradation and grey color indicates same as baseline. As can be seen, when the temperature takes different values, our proposed training scheme consistently brings performance improvement over the regular training approach on all baseline methods. Our proposed training scheme improves the overall performance. Results show that our proposed training scheme is robust in the choice of temperature, and the best result is achieved when $T=4$.

\section{Discussion and Future Work}
\label{sec:discussion}
In this paper, we proposed a novel deep learning training strategy based on knowledge distillation aiming to bridge the gap between patient-independent and patient-specific training methodologies for neurological symptom prediction. Patient-specific models, in general, perform better than patient-independent models and thus are more favored by researchers. Patient-independent models suffer from data variation across different subjects and thus often yield results not as promising as patient-specific models. Our approach takes a step from the patient-level to symptom-generating mechanisms level; we aim to teach the model to learn symptom-generating mechanisms from other patients to improve prediction performance on a specific subject. The consistent performance improvement on four state-of-the-art methods demonstrates the effectiveness of our method. 

Next, we discuss a few drawbacks of this work and potential future work. First, the output enhanced model generated with our training scheme is of the exact same size, architecture and requires the same inference time as the baseline method. However, our training scheme optimizes two models simultaneously in the second training stage and thus takes approximately twice graphics processing unit (GPU)  memory and training time, which might not be feasible to facilities with limited computation resources. One feasible solution is to prune the baseline neural network structure to reduce the parameters used or to reduce the precision of the parameter weights via quantization. However, such techniques may jeopardize the prediction performance. The trade-off between memory reduction and performance degradation needs to be carefully balanced. Second, we introduce a few hyper-parameters such as temperature, momentum, and the weight of each layer in loss function $\mathcal{L}_{dif}$. This may cause extra tuning work, thus increasing the total training time. However, through achieved experiments, our training scheme is robust to different choices of hyper-parameters. In these experiments, we used the same temperature and momentum for all methods and subjects. Besides, using gradient-based hyper-parameter tuning algorithms can also reduce the tuning time. Finally, we argue that learning symptom-generating mechanisms from other patients can improve prediction performance which is validated by experimenting with four state-of-the-art seizure prediction methods. Our experiments were conducted on the CHB-MIT sEEG database, which is a small dataset in terms of deep learning. To further improve the effectiveness of our proposed training scheme, we suggest to work with more data. In addition, we plan to cluster the latent representation vectors of data by different mechanisms in a patient-independent space. Furthermore, the clustered vectors can be projected to low-dimensional space for visualization and better interpretability.
\section{Conclusion}
\label{sec:conclusion}
We proposed a novel training scheme for better neurological symptom prediction. Specifically, our training scheme trains a pool model together with a customized model and further asks these two models to learn from each other via knowledge distillation. We demonstrate the effectiveness of our proposed training scheme by showcasing the application of seizure prediction. We adopted four state-of-the-art methods and compared performance with and without our training scheme over a benchmark dataset. Results show that the proposed training scheme consistently improves the performance of all methods in various evaluation metrics. Our training scheme can be applied to any deep learning based method with no extra modification.
\section{Acknowledgement}
\label{sec:acknowledgement}
The authors would like to acknowledge start-up funds from Westlake University to the Cutting-Edge Net of Biomedical Research and INnovation (CenBRAIN) for supporting this project. This work was supported by Zhejiang Key R\&D Program project No. 2021C03002, and Zhejiang Leading Innovative and Entrepreneur Team Introduction Program No. 2020R01005.

\section*{References}
\bibliographystyle{iopart-num}
\bibliography{ref}

\end{document}